\documentclass[letterpaper, 10 pt, conference]{ieeeconf}  

\usepackage{amssymb}
\usepackage{algorithm}
\usepackage{algpseudocode}
\usepackage{graphicx} 
\usepackage{amsmath}
\usepackage{csquotes}
\usepackage{cite}
\usepackage{rotating,booktabs}
\usepackage[verbose]{placeins} 

\IEEEoverridecommandlockouts                              
\usepackage{epsfig} 

\title{\LARGE \bf
Using RDF Summary Graph For Keyword-based Semantic Searches 
}

\author{Serkan Ayvaz$^{\ast }$,  Mehmet Aydar$^{+ }$\\
$^{\ast }$Department of Software Engineering, Bahcesehir University \\Besiktas 34353, Istanbul, Turkey \\serkan.ayvaz@eng.bau.edu.tr\\
$^{+}$Department of Computer Science, Kent State University \\Kent, OH, USA \\maydar@kent.edu
}

\begin{document}

\maketitle
\thispagestyle{empty}
\pagestyle{empty}

\begin{abstract}
The Semantic Web began to emerge as its standards and technologies developed rapidly in the recent years. The continuing development of Semantic Web technologies has facilitated publishing explicit semantics with data on the Web in RDF data model. This study proposes a semantic search framework to support efficient keyword-based semantic search on RDF data utilizing near neighbor explorations. 
The framework augments the search results with the resources in close proximity by utilizing the entity type semantics. Along with the search results, the system generates a relevance confidence score measuring the inferred semantic relatedness of returned entities based on the degree of similarity. Furthermore, the evaluations assessing the effectiveness of the framework and the accuracy of the results are presented.

\end{abstract}


\section{Introduction}

Due to rapid growth and increased complexity of data in recent years, there are many challenges in searching and gathering meaningful information efficiently from the Web of data. Data available on the Web have a variety of formats and platforms. Although the Web owes much of its success to the search engines, classical text search techniques that the search engines depend on can be problematic in finding most relevant information. 

The Semantic Web takes a big step forward by introducing the notion of Resource Description Framework (RDF)\footnote{https://www.w3.org/TR/rdf11-concepts} as the standard data model, which provides a powerful framework to overcome some of problems with the Web today. The advantage of having data represented in an unambiguous universal data model in the Web is that it allows users, applications and intelligent agents to share and make use of data automatically.   

As the amount of linked data available in RDF increases \cite{bizer_linked_2009}, it also provides a vast global platform for semantic search opportunities. Many large data sources provide a formal query end point for precise searching on the RDF data. However, despite of RDF in handling the schema and structural level interoperability, it still does not resolve the challenges in semantic reasoning in the data layer. In fact, RDF helps focus on the meaning and usage of data rather than its representation and interoperability, which is a tremendous benefit alone. 

This paper proposes a keyword-based semantic search framework that utilizes a summary graph structure for exploration of the RDF data and provides relevant results. For efficient graph explorations, a summary graph structure \cite{ayvaz_building_2015,aydar_automatic_weight} automatically from underlying RDF data is constructed.

The data retrieval tasks in Semantic Search can be roughly classified into three categorizes; Entity Search, List Search and Question-Answering. According to \cite{pound2010ad}, approximately 70\% of queries involve a semantic named entity and nearly 10\% of them are list search queries. Entities play an important role; more than half of all queries are entity centric queries. In entity search queries, the user is typically looking for a single real world object or its attributes.
In the list search queries, a user is seeking for a class or type of resources. In the Question Answering(QA) queries, users look for an answer to a specific question, which often requires natural language processing and interpretation of the question and then obtaining the most suitable answer for the question. 

The semantic search framework presented in this paper currently focuses on entity search and list search tasks of Semantic Search.In future work, we plan to extend the framework to handle QA queries by developing a mechanism for transforming and processing natural language queries.

\section{Framing the problem}

Data exploration and search using queries in a structural query language such as SPARQL have significant drawbacks. A major downside of formal query language-based systems is that the syntax can be extremely discouraging and inconvenient for non-technical users. Moreover, the requirement to know the underlying data schema before writing structured queries makes them very unlikely to be adopted by the general population. 
Consequently, integration of keyword-based queries into semantic searches on RDF graphs has attracted many research studies \cite{lei2006semsearch, fu_effectively_2011, tran_top-k_2009} since the keyword queries are easier to form and widely used in daily life.

While keyword-based queries are easier to form, there are some challenges that require more intelligent solutions than simple bag-of-words model to be addressed. For instance, the users might not know the precise set of keywords to form the intended query. Particularly, the users may not be aware of the technical terms that would be more suitable in keyword searches as is often the case in specialized domains. 

\begin{figure}[ht]
	\centering
	\includegraphics[height=7 cm]{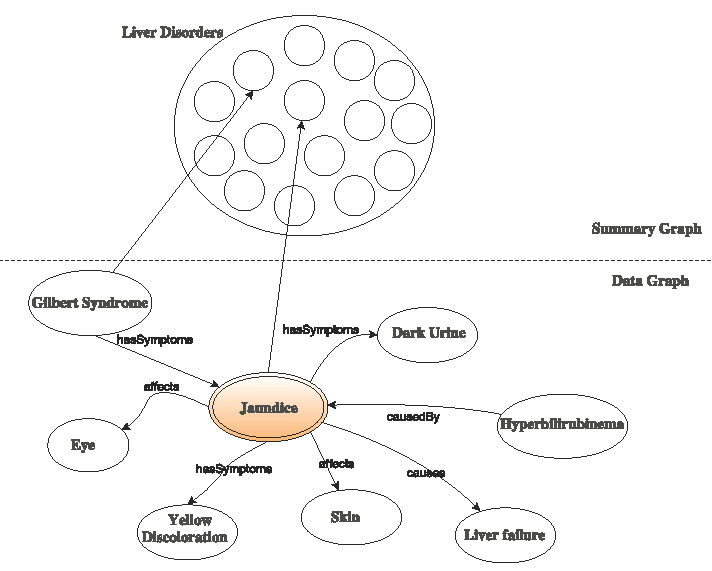}
	\caption{A Figure Matching Query Keywords to Graph Entities and the Summary Graph Nodes}
	\label{fig:figure4}
\end{figure}

Furthermore, they may not be aware of the remaining entities that might be relevant that were not part of the keyword search.  For example, let us consider a user trying to find the keyword query example searching with the keywords, ``disease causing yellow discoloration of skin and eyes'' with the intention of finding the name of a disease or related diseases based on known symptoms or pieces of information in  Figure \ref{fig:figure4}. Notice that each of these information pieces is, in fact, a feature or property describing the entity. For a user trying to find the name of a disease, it might also be beneficial to know related diseases and identify potential treatment options for the disease or disease group. 

Typically, ontological knowledge would be required to infer these semantic relations. However, the ontological knowledge may not always be available or reliable. This is due to the fact that many structured data sources available on the Web today often do not contain the type triples due to the flexibility of the RDF data model not imposing constraints on the schema. Furthermore, data publishers commonly neglect the use of standard vocabulary and describe the data using the vocabulary that is the most convenient for their purposes. Besides that, the type information can be defined too generally to be useful in identifying the entity types and related sub-types \cite{ayvaz_building_2015, aydar_automatic_weight}. Consequently, this makes it problematic for the semantic search algorithms to discover the type triples. 

The problem of keyword search in RDF graph can be formally defined as follows. Let an RDF input graph $G= (V,L,E)$ be a knowledge base such that $V$ is a finite set of resources; $E$ is the finite set of relations between resources; and  $L$ is the set of names or labels of the relations. The keyword search problem is defined as finding a set of relevant entities $v \in V$, as answers to a keyword query q.

\section{Methods}

From the Information Retrieval point of view, the result of a Web search is a set of documents on the Web containing the keywords in the user query. However, different from Web documents, an entire dataset is one large graph containing all entities related to a search over RDF data. While designing an algorithm for keyword searches over an RDF graph data, one must first determine how the results of the search should be returned. Some approaches return Steiner trees or subgraphs matching keywords in a query as answer. In our approach, we return a set of single entities as the answers. Thus, the algorithm returns a set of single entities as the result of a keyword query because we think that the users expect more granular results as answers for an entity search or a list search query. 

In this study, we consider a similar approach to \cite{tran_top-k_2009}, for keyword searches on graph-structured data based on the RDF data model, in which a summary graph structure is utilized for faster query processing. The algorithm in \cite{tran_top-k_2009} generates the top-k queries that match the user keywords using a cost-based model and allow the users to refine the structured queries. Then, the selected structured queries are processed to obtain the results.  
Differently, we do not assume that the RDF graph is well defined nor do we rely on the existence of RDF schema elements such as rdf:type rdf:class and rdf:subclass for generation of summary graph. Additionally, we do not translate the keyword queries into the SPARQL queries. Instead, we directly search the graph entities and their neighbors in the summary graph that are matching the keyword query. 

\subsection{Automatic Summarization of RDF Graph}
Data published in RDF model can be viewed as a collection of statements that intrinsically represents the resources unambiguously. An RDF statement is composed of a triple, which contains three components in form of subject-predicate-object expression that describes resources. RDF data can also be viewed as a labeled, directed multi-graph. RDF Graph nodes consist subjects and objects of triples in RDF data. A subject in an RDF triple is an RDF Internationalized Resource Identifier (IRI) reference or a blank node.  Predicates are expressed as IRIs.  An object can be either an RDF URI, a literal or a blank node \cite{cyganiak_rdf_2014}.

An RDF node representing a unique entity is expressed as an IRI node. The literal nodes in RDF data are used for values including strings, numbers, and dates. A literal node can comprise a datatype IRI, a lexical form or a language tag. 
The language tags are used only when the datatype IRI of a literal node is rdf:langString \cite{brickley_rdf_2014}. In OWL, a predicate, often called as a property of the RDF subject node, can be a Datatype property or an Object property. An Object property links a subject IRI node to an object IRI node, whereas a Datatype property links a subject IRI node to a literal object node \cite{mcguinness2004owl}. Object nodes linked to a subject node are referred as the neighbors of the subject node.

More formally, RDF graph is a directed labeled graph $G= (V,L,E)$ such that the set of vertices $V$ represent entities (resources) and entity values, the set of directed edges $E$ of the form $l(u, v)$, with  $u\in V$,  $v\in V$ and  $l \in L$, denote predicates between resources, and the labels $L$ are predicate names or labels. An edge in form of $l(u,v)$ represents the RDF triple $(u,l,v)$. 

We define a summary graph of a data graph as a directed graph such that each node in the summary graph is a subset of the original graph nodes of the same type. Thus, the summary graph generation problem is defined as obtaining the corresponding summary graph $G'= (V',L',E')$ of $G$, such that $V'$ contains equivalence classes of $V$.  $E'$ and $L'$ are, respectively, the sets of edges and labels in the graph $G'$. Hence, $L' \subset L$, and the elements of $E'$ are defined by the elements in the equivalence classes in $V'$ and the edges in $E$.

\subsubsection{Building RDF Summary Graphs}

To build a summary graph from RDF data, we compute pairwise similarity of entities. However, comparing all neighbors to all neighbors is not an efficient method in calculation of neighborhood similarity. Hence, the neighboring nodes are only compared if they are related by the same predicate. 
For instance, given two nodes $u$ and $v$, if the nodes $s_1$ and $s_2$ are neighbors of $u$, and $t$ is a neighbor of $v$. We calculate similarity of the neighborhood pairs ($s_1$, $t$) and ($s_2$, $t$) only if there is a predicate which connects $u$ to $s_1$, $u$ to $s_2$ and also connects $v$ to $t$, and we use the maximum similarity between the neighborhood pairs ($s_1$, $t$) and ($s_2$, $t$) as implied in the maximal matching concept of RoleSim similarity measure \cite{jin_axiomatic_2011}. The impact of the similarity of the neighbor nodes are weighted by each common predicate. 
For computation of similarity between two IRI nodes, the following similarity measure is used:

\setlength{\arraycolsep}{0.0em}
\begin{eqnarray}
Pa&&irSim( u,v )^{k} =  (1-\beta)  \\
&& \times  \frac{1}{|u \cup v|} \nonumber  \\
&& \times ( \sum\limits_{j \in (u \cap v)} max_{\tiny M \in Mm^{j}(u,v)} (\frac{\sum\limits_{(x,y)\in M}Sim(x,y)^{k-1} }{N^{j}_{u} + N^{j}_{v} - |M|} ) \times w_j)  \nonumber\\
&&+\beta \nonumber
\end{eqnarray}
\setlength{\arraycolsep}{3pt}

where $k$ is the iteration number, such that, if $k = 3$ then $PairSim( u,v )^{k}$ denotes to the similarity of the node pair $(u, v)$ at the third iteration and $PairSim( u,v )^{k-1}$ denotes to the similarity of the node pair $(u, v)$ by the end of the second iteration. Also, $N^{j}_{(u)}$ and $N^{j}_{(v)}$ denote their respective neighborhoods that are reached by $jth$ common edge. $x \in N^{j}_{(u)}$ and $y\in N^{j}_{(v)}$, and $N^{j}_{u}$ and $N^{j}_{v}$ denote their respective degree connected by $jth$ common edge. In other words, $N^{j}_{(u)}$ is the cardinality of $[x_{j}]$, and $N^{j}_{(v)}$ is the cardinality of $[y_{j}]$. $w_j$ is the weight of the property connecting the graph nodes $(u, v)$ and their respective neighbors $(x, y)$. 

We define $M$ to be a set of ordered pairs $(x,y)$ where $x \in N^{j}_{(u)}$ and $y \in N^{j}_(v)$ such that there does not exist $(x', y') \in M$, s.t. $x = x'$ or $y = y'$, and furthermore, $M$ is maximal in that no more ordered pairs may be added to $M$ and keep the constraint above. $Mm^{j}(u,v)$ is the set of all such $M$'s. $Mm^{j}(u,v)$ is a set of sets. 

The term ``Maximal nonrepeating matching'' represents the maximal formation of pairs from the elements in $N^{j}_{(u)}$ and $N^{j}_{(v)}$ with the restriction that no element in either $N^{j}_{(u)}$ and  $N^{j}_{(v)}$ may be used in more than one ordered pair  \cite{jin_axiomatic_2011}. The parameter $\beta$ is a decay factor.  $\beta$ diminishes the influence of neighbors with further distance due to the recursive effect and can take values in range of $0 < \beta < 1$. $l_{1}(u, x)$ and $l_{2}(v, y)$ represent directed edge labels s.t. $l_{1},l_{2}\in L,$ and $l_{1}=l_{2}$, $x \in N_{(u)}$ and  $y\in N_{(v)}$.

\begin{equation}
Sim( x, y )^{k-1}=
\begin{cases}
PairSim( x,y )^{k-1}, & \text{if} \enspace \text{x,y are IRI nodes } \\
LiteralSim( x,y ), & \text{if} \enspace \text{x,y are Literal nodes } \\
0, &  \text{otherwise }   
\end{cases}
\end{equation}

To calculate the similarity of pairs of the literal nodes, we utilize string similarities for the lexical form components. The similarity $LiteralSim( x,y )$ between two literal nodes $x$ and $y$ calculates the number of common words within the two lexical forms and it takes into account their auto-generated importance weights. When calculating the weight of word importance in literal nodes consisting of a set of words, the term frequency-inverse document frequency ($tf-idf$) \cite{luhn_statistical_1957,sparck_jones_statistical_1972}, a well-known technique in information retrieval, is used.

\subsection{Searching Over RDF Graph Data} 

In keyword-based semantic search process, two important issues to be addressed are (a) how to find a set of IRIs corresponding to each user-supplied keyword,(b) how to find related entities for given keyword query. A keyword index can be generated to store terms in RDF triples. The maching graph node hits based on exact or approximate matching can be ranked and returned as results. Finding related or associated entities is a more complicated task. For finding associated entities, we make use of the semantics obtained from the underlying summary graph. The generated summary graph structure contains the class types, relations, and entities within each class type along with a similarity matrix containing the similarity scores between entities within each graph type. The ranking of the associated entities are determined using the similarity scores in the similarity matrix. 

\begin{figure}[ht]
	\centering
	\includegraphics[height=7 cm]{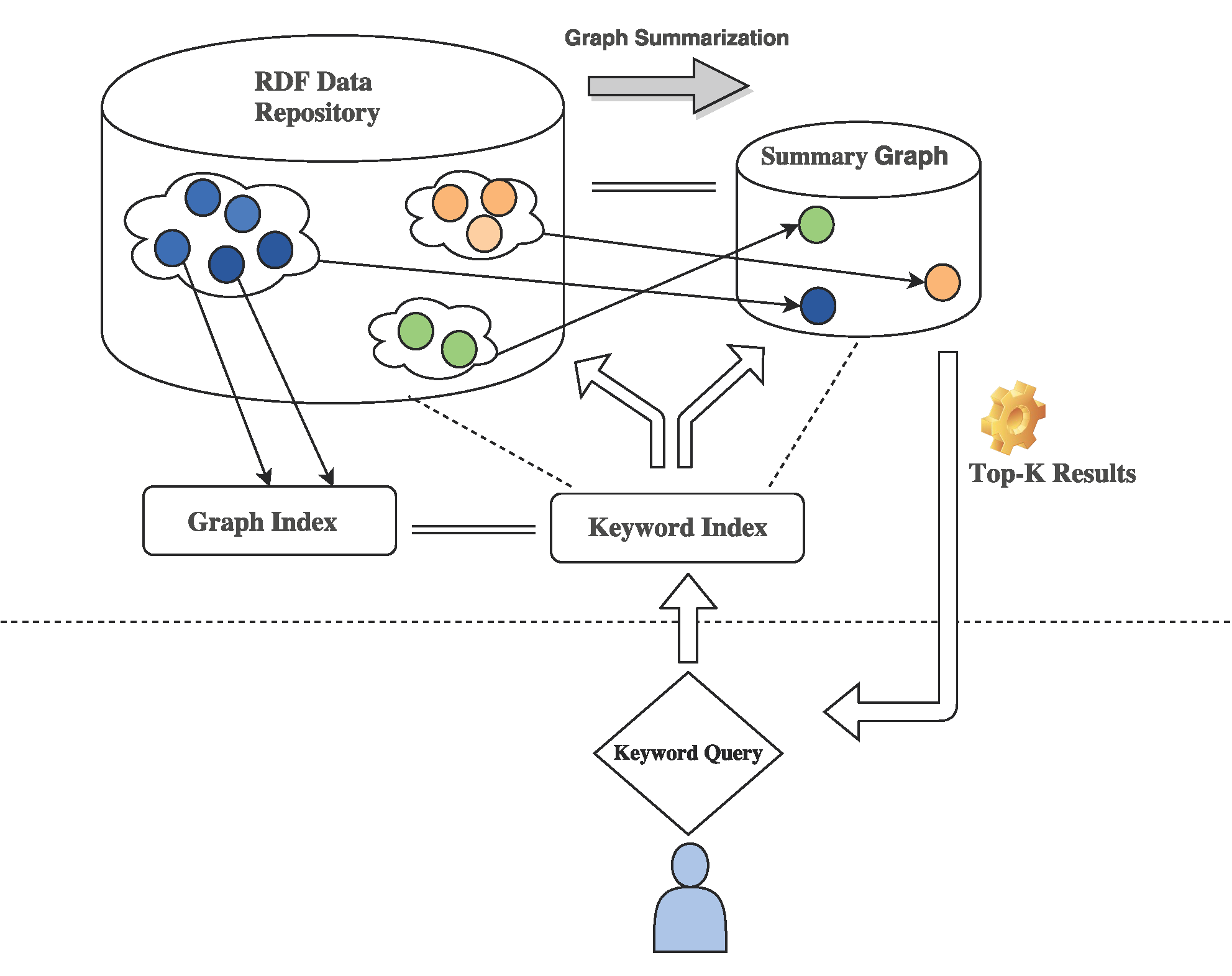}
	\caption{The System Overview}
	\label{fig:SystemOverview}
\end{figure}

The Search process involves several tasks including the summary graph generation, keyword index mapping, graph index generation which maps keywords to graph elements, ranking the results, and retrieval of the top-k relevant elements. Figure \ref{fig:SystemOverview} depicts the components and an overview of the approach.  The tasks of summary graph generation, keyword index mapping, graph index generation are performed during the preprocessing time. Hence, they do not impede the query time. Since the most of the resource-intensive computations are performed in the offline-preprocessing phase, the query processing is fast and responsive.

For tokenization, stemming and keyword indexing we use Apache Lucene\footnote{http://lucene.apache.org}, which is an open source search framework that provides efficient index technique. The keyword index maps the keyword entries to graph elements that include the entities, classes, properties or literal values. Furthermore, we supplement search candidates from the same class type in the summary graph. 

The search algorithm explores the graph nodes for input keyword query hits. Then, the neighbor nodes of the hits are considered as a candidate if their similarity is above the threshold. The candidates get ranked and top-k of them returned as the results. 

\begin{algorithm}                      
	\caption{$Search()$}          
	\algnewcommand\algorithmicinput{\textbf{Input:}}
	\algnewcommand\INPUT{\item[\algorithmicinput]}
	\algnewcommand\algorithmicoutput{\textbf{Output:}}
	\algnewcommand\OUTPUT{\item[\algorithmicoutput]}
	
	\label{Search algorithm}        
	\begin{algorithmic}                    
		\INPUT Graph G, SummaryGraph $G'$, SimMatrix S, KeywordIndex KI, GraphIndex GI
		\OUTPUT Top-K Results 
		\While {Exists(querystring)} 
		\Function{Search}{querystring} 
		\State keywords = Parse(querystring);\\
		\If {Valid(Query)=True}
		\State  hits  $ \gets  FindQueryHits(keywords)$
		\State  Nodes $ \gets  GetTopKNodes(hits) $
		\EndIf   
		\State \hspace{0.01ex}
		\Return Nodes 
		\EndFunction  
		\State  PresentResults(Nodes)     
		\EndWhile  
	\end{algorithmic}
\end{algorithm}

The ranking method takes the semantic relatedness into account when ranking the candidates. The resources with the same type and high degree of similarity to the candidate resources get better ranking scores and consequentially, elevate to a higher position in the results. We call the final relevance score a ``Relevance Confidence Score'' referring to the inferred semantic relatedness of returned entities as the answer to the keyword query. 

While providing the results of the keyword query, the results display the relevance confidence level score for the best matching candidates related to the user supplied keywords in the search. The calculated relevance scores of the answers provide insights for validation.

\begin{algorithm}                      
	\caption{$GetTopKNodes(hits)$}          
	\algnewcommand\algorithmicinput{\textbf{Input:}}
	\algnewcommand\INPUT{\item[\algorithmicinput]}
	\algnewcommand\algorithmicoutput{\textbf{Output:}}
	\algnewcommand\OUTPUT{\item[\algorithmicoutput]}  
	\label{GetTopKNodes}        
	\begin{algorithmic}                    
		\INPUT hits, Graph G, SummaryGraph $G'$, SimMatrix sim, GraphIndex index  
		\OUTPUT resultNodes
		\Function{GetTopKNodes}{hits}  
		\For{$\textbf{each} (hit \in hits)$}
		\State  Node node  $ \gets index[hit]$
		\State  \textbf{increment} $ node.hitcount$
		\State  resultNodes.Add(node);
		\State  NeighborList $\gets MatchNeighbors(Node, G')$\\
		\For{$\textbf{each} (neighbor \in NeighborList)$}
		\If{simpairs.Contains(neighbor)}
		\State neighbor.Sim $ \gets sim[neighbor,node]$
		\State \textbf{increment} $ neighbor.hitcount$
		\State resultNodes.Add(neighbor);
		\EndIf	 	 			
		\EndFor
		\EndFor  
		\State RankResults(resultNodes) \\
		\Return resultNodes
		\EndFunction    
	\end{algorithmic}
\end{algorithm}

\section{Evaluations}\label{sect:six_two} 
We evaluated the results of our keyword-based semantic search approach on a subset of DBpedia \cite{auer_dbpedia:_2007} containing 10,000 triples, with the goal of finding relevant entities answering the keyword search queries. We manually assessed the results of a set of keyword queries primarily focusing on testing two aspects of our search approach. First set of questions focused more on finding similar entities in the same class types. For example, consider a keyword query for finding entities similar to ``Acacia'', which returned the following entities along with their respective confidence scores. The relevance confidence scores depict the similarities of the resulted entities to the query.
\\*
$(http://dbpedia.org/resource/Acacia$,100\% ) \\
$(http://dbpedia.org/resource/Aloe$,71.7\%) \\
$(http://dbpedia.org/resource/Amaryllis$,71.7\%) \\
$(http://dbpedia.org/resource/Ancylopoda$,57.5\%) \\
$(http://dbpedia.org/resource/Alligatoridae$,40.5\%)\\  

Similarly in another instance, the answers to the query ``Andre Agassi'' contain the entity $Anna\_Kournikova$ in addition to the entity $Andre\_Agassi$ as follows. 
\\*
$(http://dbpedia.org/resource/Andre\_Agassi$,100\%)\\
$(http://dbpedia.org/resource/Anna\_Kournikova$,78.8\%)\\

It is important to note that neither the entity name of $Anna\_Kournikova$ nor any of its properties contain the keywords ``Andre Agassi''. The semantic similarity between these entities is inferred from the summary graph as both entities belong to the same class type in the summary graph.

The goal of second set of evaluations was identifying entities based on their descriptors rather than string matching on the entity names. The name of an entity being searched for may not always be known ahead of time. Users may want to find the name of the entity by searching on available pieces of information which are describing the entity. For instance, ``national anthem'' is a distinctive descriptor of entities that belong to the class type ``countries'' and the search results should include the entities from that class. In a similar example, the answer to the keyword query ``notable Ideas'' included the following entities since the property ``notable Ideas'' is a distinguishing descriptor for the entities that belong to the class type ``people who had influential ideas''.
\\*
$(http://dbpedia.org/resource/Aristotle)$ \\
$(http://dbpedia.org/resource/Avicenna)$\\
$(http://dbpedia.org/resource/Arthur\_Schopenhauer)$

For the assessments, we used the measures of precision, recall and F-measure, to evaluate the search results \cite{powers2011evaluation, fawcett2006introduction, _precision_2015}. Precision is defined as the ratio of correct answers over all given answers. It measures the percentage of answers that are relevant.

\begin{equation}   
Precision =  \frac{ |true\_positives| }{  |true\_positives + false\_positives|  }. 
\end{equation} 

On the other hand, recall is the ratio of correct answers over all relevant answers, which quantifies the fraction of relevant answers that are retrieved.

\begin{equation}   
Recall =  \frac{ |true\_positives| }{  |true\_positives + false\_negatives|  }.    
\end{equation}

F-Measure is the harmonic mean of precision and recall, where $0 \leqslant precision \leqslant 1 $ and $0 \leqslant recall \leqslant 1 $.

\begin{equation}   
F-Measure =  \frac{ 2*Precision*Recall  }{  Precision+Recall  }.  
\end{equation}

In summary, our framework performed with an average precision of 0.652, an average recall of 0.891 and a macro-averaged F-measure of 0.753 in the entity search set over 20 keyword queries. The accuracy of search results was manually verified. While precision was relatively lower compared to recall value, we observed high-recall (recall $>$ 0.89) in the evaluations. High-recall is crucial in search systems, particularly in specialized domains, where it is essential not to miss relevant results. Note that the accuracy of search results is dependent on the accuracy of the summary graph generation and the characteristics of the underlying dataset. Dbpedia was chosen for the evaluation as it is a well-known general purpose dataset and a good candidate for evaluating semantic search applications. Nonetheless, the performance of the framework needs to be evaluated on multiple data sources. In future work, we plan to perform further evaluations on several large datasets from various domains.
 
\section{Related Work}

RDF data have a graph structure and can be viewed as a directed graph with labeled nodes and edges. Recently, there has been a rapid increase in the amount of RDF data available on the Web, especially with the Linked Open Data initiative. This initiative also provides a vast global platform for semantic search opportunities.

Dbpedia \cite{auer_dbpedia:_2007}, FreeBase \cite{bollacker_freebase:_2008} and many other large data sources provide a formal query end point for precise searching on the RDF data. There have been many approaches, e.g. \cite{broekstra2002sesame} adopting semantic searches based on user provided queries in a formal query language such as SPARQL. Use of formal query language-based systems have potential user adoption issues as they can be difficult to use even for technical users.

On the other hand, there has been a debate on the methods to integrate keyword-based queries into semantic searches on RDF graphs since the keyword queries are easier to form and widely used in daily life. Some approaches suggested generating structured SPAQL queries based on pattern templates placing the keywords into positions in the query patterns, e.g., \cite{lei2006semsearch, shekarpour_keyword-driven_2011, Pradel_SWIP, unger2012template, ben2012medical,lehmann2011autosparql, tablan2008natural} and others proposed a cost-based exploration of the matching keywords on the RDF graph \cite{fu_effectively_2011, tran_top-k_2009}.

Some studies are solely focusing on the translation of the natural language questions into structured queries, most commonly SPARQL queries. While others go one step further by returning results for transformed queries \cite{unger2012template, ben2012medical, shekarpour_keyword-driven_2011, lei2006semsearch, Lukovnikov, hamon2014natural}.

A large quantity of these approaches based on the translation of the natural language questions into structured queries assume that some patterns or templates exist in the query keywords. They typically generate the SPARQL queries by using a parser extracting queries from Natural Language questions \cite{Pradel_SWIP} or a mechanism deriving the queries based on an ontology or knowledge base \cite{tablan2008natural}, or a supervised machine learning mechanism from Natural Language questions \cite{lehmann2011autosparql}.

There are some studies in the second group that also include the Question Answering (QA) process from beginning to end. These systems provide answers for the keyword queries. That said, many of these approaches still rely on templates or patterns in the natural language questions for transformation of queries \cite{unger2012template, shekarpour_keyword-driven_2011, Pradel_SWIP}. Some studies offer a template-based approach relying on Natural Language Processing (NLP) tools \cite{unger2012template, hamon2014natural}. In another related work \cite{ben2012medical}, a hybrid approach depending on patterns and Support Vector Machines (SVM) creates a machine learning method for extraction of the named entities and relations in the keywords for generation of the SPARQL queries. 

In our approach, we neither predefine the query patterns nor have a limit on the keyword query size. Also, we do not rely on natural language processing tools for interpretation of the keyword queries nor do we transform the keyword queries into the SPARQL queries. 

\section{Conclusion}
This article presented a keyword-based semantic search framework that utilizes a summary graph structure to enable efficient graph explorations. The system acquires the semantic type relations from the summary graph and augments the results by recommending the entities that are the same or very similar. The system utilizes the entity type information in ranking mechanism and provides relevant entities along with a relevance score, which demonstrates the semantic likelihood of returned entities as the answer to the keyword query. Additionally, the evaluations assessing the effectiveness of the framework and the accuracy of the results were presented. We observed that the framework scored a high accuracy in finding semantically related entities based on the keyword search queries.

\section*{ACKNOWLEDGMENT}
The authors would like to thank Prof. Austin Melton for his invaluable help and his guidance during the
study. This work was supported partially by Istanbul Kalkinma Ajansi Grant for Istanbul Big Data Egitim ve Arastirma Merkezi, TR10/16/YNY/0036.


\bibliographystyle{plain}
\bibliography{MainBib}

\end{document}